\documentclass[conference]{IEEEtran}
\IEEEoverridecommandlockouts
\usepackage{cite}
\usepackage{helvet}
\usepackage{courier}
\usepackage{amsmath,amsfonts,amssymb}
\usepackage{algorithmic}
\usepackage{algorithm}
\usepackage{array}
\usepackage{textcomp}
\usepackage{stfloats}
\usepackage{url}
\usepackage{verbatim}
\usepackage{graphicx}
\usepackage{epsfig}
\usepackage{booktabs}
\usepackage{makecell}
\usepackage{multirow}
\usepackage{pifont}
\usepackage{tablefootnote}
\usepackage{enumitem}

\usepackage{amsmath}

\usepackage{geometry}
\usepackage{array}

\usepackage{xcolor}

\usepackage{enumitem}

\def\BibTeX{{\rm B\kern-.05em{\sc i\kern-.025em b}\kern-.08em
    T\kern-.1667em\lower.7ex\hbox{E}\kern-.125emX}}
\begin{document}

\title{Profiling Patient Transcript Using Large Language Model Reasoning Augmentation for Alzheimer's Disease Detection 
}

\author{\IEEEauthorblockN{1\textsuperscript{st} Chin-Po Chen}
\IEEEauthorblockA{\textit{Digital Center} \\
\textit{Inventec Corporation}\\
Taipei, Taiwan \\
chen.jackcp@inventec.com}
\and
\IEEEauthorblockN{2\textsuperscript{nd} Jeng-Lin Li}
\IEEEauthorblockA{\textit{Digital Center} \\
\textit{Inventec Corporation}\\
Taipei, Taiwan \\
li.johncl@inventec.com}
}

\maketitle

\begin{abstract}
Alzheimer's disease (AD) stands as the predominant cause of dementia, characterized by a gradual decline in speech and language capabilities. Recent deep-learning advancements have facilitated automated AD detection through spontaneous speech. However, common transcript-based detection methods directly model text patterns in each utterance without a global view of the patient's linguistic characteristics, resulting in limited discriminability and interpretability. Despite the enhanced reasoning abilities of large language models (LLMs), there remains a gap in fully harnessing the reasoning ability to facilitate AD detection and model interpretation. Therefore, we propose a patient-level transcript profiling framework leveraging LLM-based reasoning augmentation to systematically elicit linguistic deficit attributes. The summarized embeddings of the attributes are integrated into an Albert model for AD detection. The framework achieves 8.51\% ACC and 8.34\% F1 improvements on the ADReSS dataset compared to the baseline without reasoning augmentation. Our further analysis shows the effectiveness of our identified linguistic deficit attributes and the potential to use LLM for AD detection interpretation. 
\end{abstract}


\section{Introduction}
Alzheimer's disease (AD) is a primary contributor to dementia, gradually impacting memory, behavior, cognitive function, and social skills. Individuals diagnosed with AD are expected to rise to over 150 million globally by 2050, three times the number in 2019~\cite{nichols2022estimation}. Early diagnosis of AD inclinations becomes increasingly crucial, but the laborious and time-consuming monitoring process using traditional assessment tools leads to a major bottleneck. As the gradually lost linguistic functions are observable indicators, automatically assessing spontaneous speech enables the tremendous into-life potential for prevalent AD detection~\cite{gomezzaragoza23_interspeech}. 
Automatic speech recognition (ASR) technologies also ensure the accessibility of spoken transcripts in clinical practice.

Recent deep learning models are applied to learn the underlying linguistic pattern from the transcripts.
The CNN and BLSTM network architectures have shown discriminative power~\cite{karlekar2018detecting}, and BERT-like models further improve AD detection with fine-tuning techniques~\cite{ilias2022explainable}.
These transcript-based detection methods involve less sensitive data than the risk of identity leakage in speech. 
However, the investigation of language deficits is limited, as past studies only concern the linguistic pattern within an utterance for feature extraction, without the viewpoint from the understanding of a whole session. For example, the identified local low-level features such as pauses and punctuation can only characterize the deficits in an influent spoken utterance~\cite{gomezzaragoza23_interspeech}. 





Local low-level feature representation constrains the modeling ability for patient-level AD detection tasks, which biases the predictive models and limits the explainability. There is a research gap in generating global high-level representations that systematically summarize the sessional-level narrative. Fortunately, recent pre-trained large language models (LLMs) have demonstrated remarkable reasoning ability for summarization based on user-given prompts. A recent study~\cite{wang2023exploiting} finetunes an LLM with prompt templates such as ``diagnosis is $<$MASK$>$'', where $<$MASK$>$ can be either ``dementia'' or ``healthy''. However, the simple prompt design makes no difference to the standard classification task, i.e., not fully leveraging the reasoning ability of LLMs.

This study proposes a patient-level transcript profiling framework using LLM-based reasoning augmentation to systematically enhance AD detection. We design an LLM prompt to extract and summarize multiple linguistic deficit attributes over each patient dialogue session. Our framework can thus generate a patient-level profile to augment the discriminative ability. Our evaluation of the reasoning-augmented AD detection framework achieves an 8.34\% improvement, compared to the model without patient profiling on the ADReSS Challenge dataset. Our analysis reveals that the model training is affected by healthy control participants generating linguistic deficits. We also provide a case study for the patient transcript profile to illustrate the enhanced model explainability for clinical applications.

\begin{figure*}[t]
\centerline{
  \includegraphics[width=0.95\textwidth]{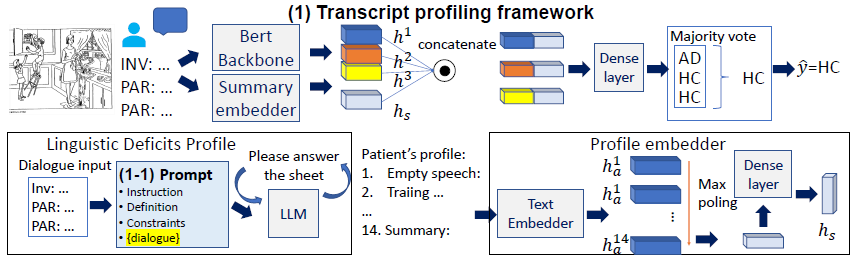}
  \vspace{-2mm}
}
\caption{
{\bf Overview of our proposed {\em LLM reasoning augmented} AD detection framework.} The transcripts from each participant and a designed prompt are used for linguistic attribute detection and summary using an LLM. The attributes and summary are transformed as embeddings to concatenate with the latent layer of the BERT backbone model for AD and HC sentence classification. Finally, for each participant's session, we conducted a majority vote on the prediction results of each sentence from the dialogue and form the final prediction $\hat{y}$.}
\label{fig:framework}
\vspace{-5mm}
\end{figure*}

\section{Methods}
\vspace{-2mm}
\subsection{Dataset and Study Population}

We study the AD detection system using the ADReSS Challenge dataset~\cite{luz20_interspeech} which includes 156 American participants with gender and age balanced. Each participant (Par) was asked to describe the events in the Cookie Theft by an investigator (Inv), which is an assessment from the Boston Diagnostic Aphasia Exam~\cite{becker1994natural}. Recorded audio data are manually transcribed as textual data. The healthy control (HC) and AD groups both contain 54 participants for training and 24 participants for testing.



\begin{figure*}[b]
\centerline{
  \includegraphics[width=\textwidth]{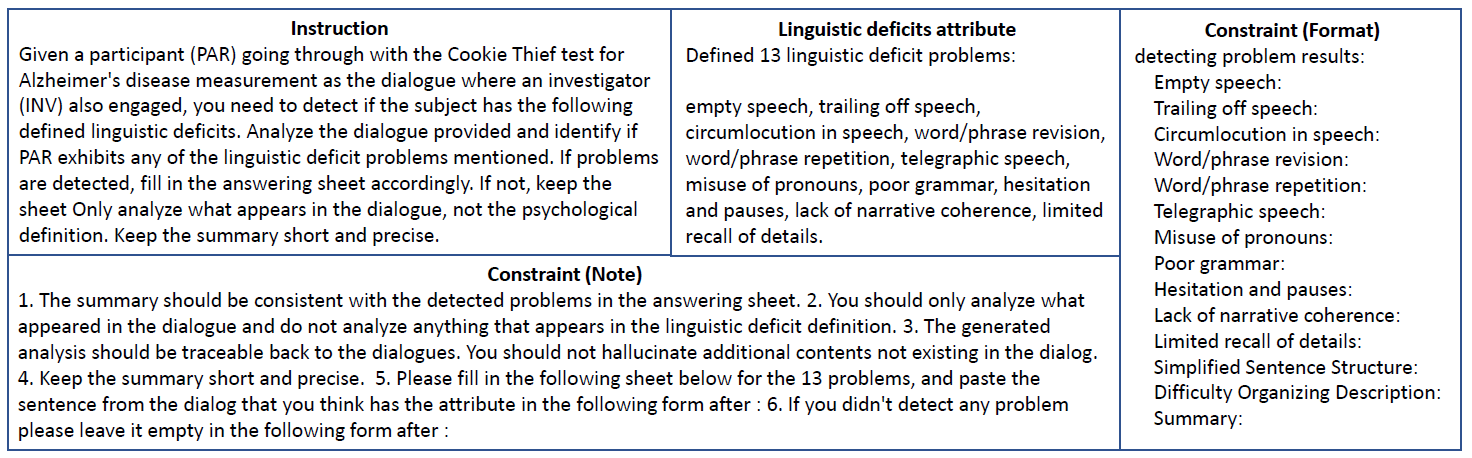}
  \vspace{-2mm}
}
\caption{
{\bf Prompt designed to instruct LLM for personal linguistic profiling.} 
}
\label{fig:prompt}
\vspace{-5mm}
\end{figure*}

\subsection{Reasoning Augmented BERT}
\label{ssec:RABERT}

We designed a reasoning scheme based on 13 linguistic deficit attributes ($\S$\ref{sssec:proxy_attribute}) as a proxy to generate the participants' linguistic deficit profile and then embed that into the BERT-based model for AD detection ($\S$\ref{sssec:model}).

\subsubsection{Linguistic Deficit Profile}
\label{sssec:proxy_attribute}
To generate a linguistic deficit profile, our designed prompt for LLM, shown in Fig.~\ref{fig:prompt}, contains four parts: instruction, linguistic deficits attribute description, notification constraints, and format constraints. These parts are designed to secure profile quality by constraining the output format of the LLM. 

The second part, linguistic deficits attribute description, specifically emphasizes the categories of clinically relevant information.
Language deficits of Alzheimer's patients have been identified as a few measurable tasks, including anomia, dysfluency, and agrammatism in the previous work~\cite{gkoumas2023reformulating}. 
However, these human-defined attributes might not be comprehensive enough to generate a linguistic deficit profile. In this circumstance, we introduce 13 linguistic deficit attributes by extending these measurable tasks and querying the LLM for a refined definition of the attributes. The derived attributes contain empty speech, trailing off speech, circumlocution in speech, word/phrase revision, word/phrase repetition, telegraphic speech, misuse of pronouns, poor grammar, hesitation and pauses, lack of narrative coherence, and limited recall of details. The details are provided in \footnote{\url{https://github.com/JackingChen/Reason_Augment_AD_detect.git}.}.

We utilize gpt-35-turbo engine as our LLM provided by Azure OpenAI, a well-established and easily accessible ChatBot. To stabilize the response, we submit a follow-up prompt `Please answer the sheet' along with the first-turn response to derive the final patient linguistic profile. We leverage text-embedding-ada-002 as a text embedder to derive 1536-dimensional attribute embeddings $h_a^{i} \in \mathbb{R}^{1536}$, where $i\in \{1...14\}$ for a varied number of detected attributes and a summary for each participant. Then, a patient profile embedding $h_s\in \mathbb{R}^{d_s}$ with $d_s$=512 is derived by performing max-pooling on these attribute embeddings followed by a dense layer. This operation picks up the most salient attribute across all profile embeddings and turns them into a more compact feature representation.

\subsubsection{Augmented BERT for AD Detection}
\label{sssec:model}
We use Albert~\cite{lan2019albert} as a backbone to process tokenized textual inputs from transcripts. The Albert network is an efficient BERT-like model featuring the enhanced ability to handle long paragraphs. Figure~\ref{fig:framework} shows our proposed framework. Each dialogue contains a varying number of sentences $T$, serving as inputs to the BERT backbone. For each BERT latent embedding $h^{j}\in \mathbb{R}^{d_h}$ with $d_h$=768 and $j\in\{1...T\}$, we concatenate the participant's profile embedding $h_s$ to obtain $h_{concat}=h^{j} \odot h_s$, which augments the feature space in a personal profile-aware manner. Then, a two-layer dense network predicts sentence-level AD results, which are aggregated by a majority vote for the final prediction of a participant. 
\subsubsection{Model configuration}
The BERT backbone's configuration is the same as in HuggingFace\footnote{https://huggingface.co/}. The dimensions of the two-layer dense layer are 640 and 2, and that of the dense layer in the profile embedder is 512. In addition, we use the AdamW optimizer with a learning rate of 2e-5, and train the proposed model with 4 epochs. 

\begin{table}[htbp]
    \centering
    \caption{AD detection results on different evaluation metrics in percentage (\%). RA denotes reasoning augmentation described in \ref{sssec:proxy_attribute}.}\vspace{-3mm}
    \resizebox{\columnwidth}{!}{%
    \begin{tabular}{c|c|c|c|c}
        \toprule
        Model & Precision & Recall & Accuracy & F1 \\
        \midrule
        Baseline: Albert & 72.75 & 68.25 & 70.83 & 69.28 \\
        \hline
        \hline
        Mbert & 76.39 & 70.11 & 72.92 & 71.16 \\
        \hline
        ClinicalBERT & 72.89 & 65.34 & 68.75 & 65.85 \\
        \hline
        Mbert+RA$_{3}$ & 74.69 & 67.72 & 70.83 & 68.55 \\
        \hline
        Mbert+RA$_{13}$ & 74.58 & 70.63 & 72.92 & 71.76 \\
        \hline
        Albert+RA$_{3}$ & 74.60 & 74.60 & 75.00 & 75.00 \\
        \hline
        Albert+RA$_{13}$ & \textbf{79.89} & \textbf{77.78} & \textbf{79.17} & \textbf{78.79} \\ 
        \bottomrule
    \end{tabular}%
    }\vspace{-3mm}
    \label{tab:performance_metrics_resized}
\end{table}


\subsection{Experimental Analysis}
\subsubsection{Experiment I: Detection Results Comparision Using Different Language Models}
We use four metrics, including precision, recall, accuracy, and F1 score, to evaluate model performances. The compared models are listed:
\begin{itemize}[leftmargin=10pt] \itemsep -.1em
    \item BERT: BERT-like models such as Albert~\cite{lan2019albert}, Mbert~\cite{mbert2018}, or ClinicalBERT~\cite{clinicalbert2019} which are used in the previous works~\cite{ilias2022explainable}.
    \item BERT+RA$_{3}$: a baseline using the three proposed attributes (anomia, dysfluency, agrammatism) from~\cite{gkoumas2023reformulating} for LLM reasoning augmentation.
    \item BERT+RA$_{13}$: our proposed method using the LLM reasoning augmentation described in $\S$\ref{ssec:RABERT}.

\end{itemize}

\subsubsection{Experiment II: Investigation of Effects on Patient Profiling}
\label{sssec:exp2}
We conducted an analysis and a case study to examine the effects of profiling the patients' transcripts. 
Our analysis aims to investigate the effects of incorporating patient profiles in model training. We first define a risk-ascend index to indicate the increment of the model's tendency to classify a participant as AD. This is calculated by subtracting $P$, the percentage of sentences predicted as AD, between our proposed model (Albert+RA$_{13}$) and the baseline model (Albert): $\delta=P^{\hat{y}=1}_{proposed}-P^{\hat{y}=AD}_{baseline}$. We thus derive $\overline{\delta}_{hc}$ and $\overline{\delta}_{ad}$ by averaging over the utterances generated by the HC and TD groups. Greater $\overline{\delta}$ indicates our model regards more utterances of a person to be likely generated by AD than the baseline model, and vice versa. Since this index indicates the models' decision change, it is used to identify the contributing reasons for the improved accuracy of our model.

The participant selected for our case study is based on the observation of the analysis. We consider an HC participant with several detected attributes and a high risk-ascend index. Based on the observation from the case study we qualitatively discuss the impact of similar kinds of samples to the model. 

\begin{table}[]
\caption{Result of the average risk-ascend index ($\overline{\delta_{hc}}, \overline{\delta_{ad}}$) of HC and AD groups. $N_{attr}$: number of detect attribute; $N_{hc}$, $N_{ad}$: number of HC and AD participants.$\hat{y}_{hc}$, $\hat{y}_{ad}$ number of correctly predicted HC and AD participants}\vspace{-3mm}
\label{tab:attr_statistics}
\resizebox{\columnwidth}{!}{%
\begin{tabular}{l|lll|lll}
\hline
$N_{attr}$ & $N_{hc}$ &  $\hat{y}_{hc}$  & $\overline{\delta}_{hc}$ & $N_{ad}$ &  $\hat{y}_{ad}$  & $\overline{\delta}_{ad}$ \\ \hline
1          & 13       & 12 & 17.1                 & 21       & 14 & -3.1                 \\
2          & 4        & 3 & 15.3                 & 7        & 6 & -1.3                 \\
3          & 1        & 1 & 10                   & 2        & 2 & 0                    \\ \hline
\end{tabular}%
}\vspace{-3mm}
\end{table}

\section{Results}
\noindent\textbf{Exp I}:
The comparison among different AD detection methods is reported in Table~\ref{tab:performance_metrics_resized}. Our proposed Albert+RA$_{13}$ achieves a comprehensive relative improvement over several evaluation metrics, e.g., 7.14\% precision, 9.53\% recall, 8.34\% accuracy, and 8.51\% F1 against the Albert baseline. The compared backbone networks, including Albert, Mbert, and ClinicalBERT, attain 70.83\%, 72.92\%, and 68.75\% in accuracy, respectively. The Mbert network with multilingual pre-trained achieves the highest prediction score among the baselines. Even though ClinicalBERT is trained on clinical notes, the domain gap between the original training data and the current task still limits its performance.


Our proposed approach,
Albert+RA$_{13}$ outperforms Mbert+RA$_{13}$ with 6.25\% in accuracy and 7.03\% in F1. Albert, equipped with an efficient network design in contrast to Mbert, alleviates overfitting when incorporating the profile embedding. 
Moreover, Albert+RA$_{13}$ surpasses Albert+RA$_{3}$ 4.17\% in accuracy, implying the enhanced profile quality of our proposed linguistic deficit attributes. The additional attributes include cognitive-related aspects such as limited recall of details, which enriches the detection space at a higher level. 


\noindent\textbf{Exp II}:
Table~\ref{tab:attr_statistics} shows the risk-ascend index of the Albert+RA$_{13}$ for both HC and AD groups. We report the corresponding number of participants with their detected number of attributes. 
We found that the HC group with detected language deficit attributes tends to obtain a large positive risk-ascend index ($\delta_{\text{hc}}>10$), and the AD group usually obtains small negative values of the index ($-3<\delta_{\text{hc}}<0$). According to the result, our proposed model regards more utterances from these HC participants as utterances from AD. The prediction changes of these utterances altered the decision boundary, causing the reasoning augmented model to have better results.


We explore several cases and report one HC participant (S018) in Table~\ref{tab:case_study} as an example to illustrate our findings. Based on the previous analysis, the proposed model became less confident in those HC samples with attributes detected. We further investigate these cases and observed that these HC participants deliver more diverse and detailed content in the picture. This behavior involves more complex cognitive processes like creative thinking than the cognitive effort needed in simply describing what's in the picture. For example, many hesitations and pauses were detected when the participant started to discuss whether it was snowing out of the window even if there was no sign of snow in the picture. The whole story lacks coherence because he interrupts the original descriptions with sentences like ``I don't know''. Moreover, the LLM then considers ``I don't know'' as a sign of limited recall of details.

\begin{table}[]
\caption{Detected attributes in the profile of participant `S018'}
\label{tab:case_study}\vspace{-3mm}
\resizebox{\columnwidth}{!}{%
\begin{tabular}{l|l}
\hline
\begin{tabular}[c]{@{}l@{}}Hesitation \\ and \\ pauses\end{tabular}     & \begin{tabular}[c]{@{}l@{}}Examples:\\ "UH JUST GO AHEAD AND TELL   YOU"\\ "UH SHOES ARE GETTING WET IN THE MOTHER"\\ "I DON'T KNOW\end{tabular} \\ \hline
\begin{tabular}[c]{@{}l@{}}Lack of \\ narrative coherence\end{tabular}  & \begin{tabular}[c]{@{}l@{}}Description:\\ The descriptions seem disjointed and fragmented.\end{tabular}                                          \\ \hline
\begin{tabular}[c]{@{}l@{}}Limited \\ recall of \\ details\end{tabular} & \begin{tabular}[c]{@{}l@{}}Examples:\\ "I DON'T KNOW THAT SNOW IS   ACTION"\\ "I DON'T SEE IT SNOWING"\end{tabular}                              \\ \hline
\end{tabular}%
}\vspace{-3mm}
\end{table}


\section{Discussions}
Our results show encouraging improvements using linguistic deficit profiling on automatic AD detection. Our proposed personal profiling framework can augment the BERT-like AD detection perspective. 
The LLM's reasoning capability might provide explainable attributes to the clinicians. We plan to involve clinicians to improve these attributes and anticipate additional insights generated in a human-machine collaboration paradigm. Our analysis illustrates the current overconfident prediction of HC participants who exhibit natural linguistic deficits. It will be intriguing to validate the identified linguistic deficits as an early-detection indicator for HC participants with AD risks in the future.   
Speaking of real-world deployment, transcript-based AD detection can alleviate sensitive information without suffering the risks of revealing patients' identities in speech cues. 

\section{Conclusions}
This study demonstrates the potential of leveraging LLMs to assist in detecting Alzheimer's disease. We introduce the reasoning augmentation framework to enhance the discriminability and explainability of BERT-based language models. We anticipate that systematically profiling linguistic behaviors can facilitate early diagnosis and disease status monitoring for current and potential Alzheimer's patients. The framework can extended to handle other neurocognitive and mental disease assessments, such as depression, autism spectrum disorder, and post-traumatic stress disorder.

\bibliographystyle{IEEEtran}
\bibliography{dementia_llama}


\end{document}